\documentclass{article}

\usepackage{PRIMEarxiv}

\usepackage[utf8]{inputenc} 
\usepackage[T1]{fontenc}    
\usepackage{hyperref}       
\usepackage{url}            
\usepackage{booktabs}       
\usepackage{amsfonts}       
\usepackage{nicefrac}       
\usepackage{microtype}      
\usepackage{lipsum}
\usepackage{fancyhdr}       
\usepackage{graphicx}       
\graphicspath{{media/}}     
\usepackage{enumitem}
\usepackage{natbib}

\title{Agency in the Age of AI
}

\author{
  Samarth Swarup \\
  Biocomplexity Institute \\
  University of Virginia \\
  Charlottesville, VA\\
  \texttt{swarup@virginia.edu} 
}

\begin{document}
\maketitle

\begin{abstract}
There is significant concern about the impact of generative AI on society. Modern AI tools are capable of generating ever more
realistic text, images, and videos, and functional code, from minimal prompts. Accompanying this rise in ability and usability,
there is increasing alarm about the misuses to which these tools can be put, and the intentional and unintentional harms to
individuals and society that may result.
In this paper, we argue that \emph{agency} is the appropriate lens to study these harms
and benefits, but that doing so will require advancement in the theory of agency, and
advancement in how this theory is applied in (agent-based) models.
\end{abstract}


\section{The Problem}
\label{sec:problem}


We are living in a time of rapid and far-reaching change, the ``Age of AI'', driven by emerging AI technologies and tools. This is naturally 
giving rise to a range of concerns about the impacts of these tools, particularly generative AI tools, on society. Our goal, and blue sky idea, 
is to lay out a research perspective for addressing these concerns, especially to mitigate the potential harms. We begin by summarizing 
the problem, before laying out the theoretical and methodological advancements we imagine for addressing it. 

There are multiple kinds of harms from generative AI that have been hypothesized and discussed in the recent literature. These can be grouped 
into the following categories:

\smallskip
\noindent
\textbf{Misuse by malicious actors:} A commonly discussed example is, emerging threats to democracy. \cite{Kreps2023democracy} summarize the three main 
challenges as threats to representation, accountability, and trust. Lawmakers (elected representatives) and 
rule-making agencies respond to public concerns, which are expressed through messages sent to them by the public, by making new laws and 
regulations. This process can be subverted by malicious actors through the use of AI tools by generating large numbers of biased or fake 
messages. Similarly, accountability, i.e., the ability of the public to choose their representation through fair elections, can be subverted by, 
e.g., the spread of misinformation to influence opinions on social media. The repeated misuse of these tools, or even the perception of misuse
can lead to the erosion of trust in the democratic process. Other examples of misuse by malicious actors include the spread of misinformation 
during crises such as epidemics~\citep{Blauth2022AIcrime}, leading to inefficient decision-making and implementation of interventions, and the 
facilitation of cyberterrorism by the increased ability to find exploits~\citep{Bernardi2024}, making people and organizations more vulnerable.

\smallskip
\noindent
\textbf{Exploitation of the changed information landscape:} Just the presence of generative AI tools and their possible use creates some
pernicious effects. The main example here is the ``liar's dividend'', where, e.g., a politician or a company can claim that any evidence of 
wrongdoing by them is fake (created by their opponents using generative AI tools)~\citep{Schiff24liar}. This effectively makes it hard for the 
public to judge evidence and to properly evaluate their choices when voting, buying products, etc.

\smallskip
\noindent
\textbf{Corruption of the tools themselves:} As these tools are being trained and iteratively improved, they can also be prone to corruption. 
Attackers can try to manipulate ML models by, e.g., providing bad training data intended to bias the model or reduce its trustworthiness. These 
are known as \emph{integrity attacks}~\citep{Blauth2022AIcrime}. An unintentional version of this is the ``institutionalization of 
misinformation'', where spurious information generated by Large Language Models (LLMs), say, is used to train other LLMs iteratively, until it 
becomes firmly entrenched or institutionalized~\citep{Garry2024llm}. This is an instance of a broader class of algorithmic bias issues, which 
have many types and sources~\citep{Fazelpour2021bias}. While the commonly known and used AI tools are being built with attempts at safeguards, 
malicious actors could also make versions of these tools which are intentionally designed to function in biased or otherwise pernicious ways. 
Deterioration in the quality of these tools can exacerbate both algorithmic bias and automation bias (as described below).

\smallskip
\noindent
\textbf{Unintended consequences:} Even without malice, generative AI tools are having such a transformative impact on society that unintended 
consequences are inevitable. An example is the rapid changes occurring in the job market, such as a sharp reduction in software engineering jobs 
due to increasing use of AI tools for coding~\citep{kessler24coding}. Other more indirect examples include automation bias and selective 
adherence to algorithmic advice~\citep{AlonBarkat2022}. Automation bias refers to the uncritical use of advice from AI decision support
systems, which is problematic because these systems are not perfect and have inbuilt and poorly understood biases of their own. On the flip
side, when human decision-makers selectively adhere to advice from AI systems, there is the risk that they might do so only when the advice 
supports preconceived notions and stereotypes. The larger and more subtle societal risk here is the gradual loss of control to AI 
decision-makers, whereby humanity might gradually cede control over our society and future~\citep{Bernardi2024}. 

The range of harms, thus, is widely varied, and the listing and categories above are by no means comprehensive. An important question at this 
stage is, do we have to deal with these problems piecemeal, as we encounter them, or can we develop a theoretical and methodological framework 
that will give us a common perspective on these problems?

\section{The Idea}
\label{sec:idea}

We believe that \emph{agency} is the most appropriate theoretical lens to view the problems outlined above. There are multiple
theories and definitions of agency~\cite[e.g.,][]{kiser99agency, shapiro05agency, Hodges2022agency, emirbayer98agency, Tasselli2021}.
The most commonly accepted theory in the multi-agent systems community is the Planning Theory of Agency, due to~\citet{Bratman1987intentions, 
bratman2007structures, bratman13shared_agency}. Broadly, autonomous agency is the ability to choose your own goals and make effective plans to 
achieve them, i.e., ``the agent herself directs and governs her practical thoughts and action''~\cite[p.4]{bratman2007structures}. This theory 
has been computationally operationalized as the Belief-Desire-Intention (BDI) model~\citep{rao1995bdi}, and its many variants.

Briefly, a BDI agent has \emph{beliefs} about (the current state of) the world, \emph{desires} for preferred states of the world, from which
it selects a \emph{goal}, and \emph{intentions} or (partial) plans about how to achieve its goals. This model has been tremendously 
successful in driving research in MAS. To see the usefulness (and limitations) of this model in unifying the problems from the previous
section, let us briefly take an adversarial perspective.

Consider a thought experiment with an agent and an adversary. The agent has its beliefs, goals, plans, etc. The adversary's goal is to limit or 
reduce the agent's agency. What are the ways in which it could do this? We refer to the following as the adversary's attacks on agency.
\begin{enumerate}[label=(A\arabic*)]
    \item It could counteract the agent's plan by preventing its actions from succeeding somehow.
    \item It could try to affect the agent's planning process so that it fails to come up with effective plans.
    \item It could try to influence the agent's desires and goal selection, so that the agent only develops desires and goals that align with the adversary's choices.
    \item It could try to influence the agent's beliefs so that the agent thinks its desires are unachievable.
    \item It could try to affect the agent's belief formation system, so that the agent is unable to develop
    true beliefs by, e.g., providing misinformation or affecting the agent's ability to distinguish good and bad information.
    \item It could change the environment such that the agent only has bad options available.
\end{enumerate}

Let us consider the problems from the previous section in light of this approach. The threats to representation and accountability in
democracy fall under attacks A3, A4, and A5, as does the spread of misinformation during crises. The threat of trust erosion and that of 
increased cyberterrorism falls under A6. The liar's dividend example is an instance of attack A5, as it affects the agent's ability to 
distinguish true and false information.

Tools extend agency by extending our cognitive abilities and by extending the range of actions we can take, thereby allowing new desires
and correspondingly new plans. Thus, corruption of the tools results in a reduction of agency in a variety of ways. The specifics of the ways 
the tools are used and how they are corrupted determine the class of  attack on agency. For example, AI tools can be used in planning (one's 
schedule, one's investment  strategy, etc.), and their corruption in this context would be an instance of A2. 

Clearly, most of these require that the adversary have much more control and (computational) power than the agent, which has made it impractical
in the past for large-scale attacks on the agency of individuals and populations. However, this is precisely what has changed with the
advent of generative AI tools. These tools make it relatively trivial to generate large volumes of customized messages, which can be used
to power bot armies online, for example for attacks A3--A5. The more widespread these attacks become, the more the other attacks become possible 
as well.

Finally, for the set of problems classified as unintended consequences, we note that the above attacks can happen unintentionally as well.
For example, if a person enrolls for a programming certificate course, thinking that they will easily get a software job at the end of it,
only to find that there has been a sharp drop in the demand for such jobs due to the emergence of AI coding tools~\citep{kessler24coding}, this
is effectively an instance of A6, albeit non-adversarial. Automation bias and selective adherence result in suboptimal planning and thus
are instances of A2, and the loss of control to AI decision-makers is an extreme instance of A3.

We see, therefore, that the general conceptual framework of the Planning Theory and the BDI model bring together most potential harms from 
modern AI tools into a common set of underlying principles. However, in doing so, they also expose a number of limitations of the BDI model in 
particular, and of the Planning Theory more broadly. From the above discussion, we summarize these as the following observations:
\begin{enumerate}[nolistsep,label=(O\arabic*)]
    \item \textbf{Agency is fundamentally a multi-agent phenomenon.} It is not that we cannot speak of the agency of a single agent but, as we 
    see above, we need a multi-agent setting to see the range of phenomena involved that need to be part of a more complete theory of agency. In 
    reality, this picture is even more complicated as agents can be overlapping groups of individuals, organizations, etc., any of which can be 
    allies or adversaries of each other. While the Planning Theory has been extended to small groups~\citep{bratman13shared_agency}, neither it 
    nor the BDI model are yet capable of handling the general case.
    \item \textbf{Agent cognition needs to be a part of a theory of agency.} Attacks A2--A5 are aimed at the agent's mental state and/or the 
    process by which the agent perceives, reasons, and plans, i.e., on the cognitive mechanisms of the agent. These are not typically
    considered intrinsic to the Planning Theory or the BDI model, but should be in this perspective.
    \item \textbf{Self-monitoring needs to be part of agent cognition}, so that the agent can assess its own agency. If it cannot detect
    attacks on its agency, it effectively doesn't have agency since, e.g., it would continue to re-plan despite every plan failing due to
    the adversary's actions.
    \item \textbf{A theory of agency needs to be quantitative} in its assessment of an agent's agency, to make it
    possible to understand if an agent's agency is being diminished (or augmented). This is not to say we need a single
    number to quantify agency, since it doesn't make sense to compare agency across domains (does a doctor have more agency than a lawyer? It 
    depends on whether they are dealing with a medical issue or a legal issue), but we do need the ability to quantify the impact of attacks
    on agency.
\end{enumerate}

There are multiple strands of research which need to be brought together to create this extended theory and model. We discuss these next,
along with the challenges in incorporating these advancements into agent-based models and simulations.
\section{The Approach}
\label{sec:approach}

Different strands of research have developed different and complementary pieces of the blue sky theory we envision. These have to be
brought together, extended, and integrated into a cohesive whole and then operationalized into a model. We summarize these strands below.

In keeping with observation O1, the sociological Relational Theory of Agency~\citep{emirbayer98agency, burkitt16relational} gives primacy to the
interactions among agents. Their definition of agency is surprisingly similar to the Planning Theory, with some key differences. In their view, human agency
consists of three parts~\citep{emirbayer98agency}. The first is an ``iterational element'', which consists of ``the selective reactivation of 
past patterns of thought and action'', which is very similar to a plan library. The second is a ``projective element'', which consists of ``the
generation of possible future trajectories of actions'', i.e., the generation of (partial) plans. The third is a ``practical-evaluative 
element'', which is the capacity ``to make practical and normative judgments among alternative possible trajectories of action''. The key
differences are that this perspective emphasizes social interactions (in the practical-evaluative element) and the cognitive aspects of planning
(in the iterational and projective elements). A limitation is that the social interactions considered are not adversarial in particular, thus 
extra work is needed to apply it to our problems of interest. While this theory is not mathematically or computationally formalized, perhaps it 
would not be a big step to do so.

More cognitively-oriented theories of agency, which are aligned with O2 and O3, are based on the ``enactive'' perspective. Enactive agency 
emerges from the interactions among agents~\citep{dejaegher09socialInteraction} and the interactions between agents and their 
environments~\citep{dipaolo17sensorimotor}. However, this theory is more focused on sensorimotor interactions, with a view to defining the
minimal conditions for agency. This contingent nature of agency has been recognized in the AAMAS community from its early days as 
well, when Franklin and Graesser pointed out that agency has be defined with respect to an environment~\citep{franklin97agentTaxonomy}.
In the enactive view, a system is an agent if it meets three requirements: self-individuation, interactional asymmetry, and 
normativity~\cite[Ch. 5]{dipaolo17sensorimotor}. Self-individuation means that the agent must be able to distinguish itself from its 
surroundings. Interactional asymmetry means that the agent is capable of initiating actions. Normativity means that these actions are
performed according to the agent's goals and norms. Thus, once again, there is a notion of goals and (implicitly) plans here, but embedded
in a social and environmental context. The added requirement is self-individuation, which has largely been taken for granted in the Planning
Theory and the BDI model.

However, defining a boundary between the agent and its environment can be tricky as agents can extend their cognition into the
environment~\cite[e.g.,][]{williams10infoDynamics}, and agency can be socially distributed in other ways~\citep{gasser91social}, including
over social networks~\citep{Tasselli2021, Mehrab2024}. But, we believe that this is a very important piece of the puzzle for the problems
we have set out to address. Simply put, if some of the cognition of a person takes place using environmental resources (e.g., computers),
this makes it vulnerable to attacks via those resources (e.g., computers can be hacked and generative AI tools are making it easier).

The use of information theory is promising in this regard as it might also offer a means of measuring the agency of an agent in an 
environment~\citep{jung11empowerment}, thus addressing O4. More recent work along these lines proposes the use of Markov 
blankets~\citep{ramstead2021multiscale} or causal blankets~\citep{rosas20causal}. In fact, Ramstead et al.'s multiscale perspective is that
the boundaries of cognition are dynamically maintained across multiple spatiotemporal scales~\citep{ramstead2021multiscale}. In their 
formulation, organisms act to minimize surprise, which turns out to be equivalent to minimizing a variational free energy. This is similar to
the empowerment idea~\citep{jung11empowerment}, and perhaps a generalization of it. The main limitation is that it is hard to apply this
formalism practically to any but the simplest of agents and environments.

Bringing it all together, our vision of a new theory of agency combines the relational and cognitive perspectives, while still including
the idea of beliefs, goals, and partial plans that are the necessary components of the Planning Theory. This would be a significant
extension of that theory, and its operationalization would be a significant extension of the BDI model. Further, we envision, 
information-theoretic ideas would be used to make this theory quantitative in the sense of providing a framework to judge how much agency changes
dynamically due to the interactions among agents, adversaries, and the environment. In practical terms, to make these judgments, we would
turn to agent-based models and simulations, so that these interactions can be evaluated in particular scenarios.

\subsection{Agent-based Modeling and Simulation}

While agent-based models (ABMs) are used in many domains, we are not aware of their use yet in modeling the harms due to generative AI
tools. LLMs are increasingly being used in ABMs to model human behavior, but they are not being used as LLMs qua LLMs~\citep{Gao2024llm-abm}.
Our idea is to develop ABMs that include representations of ``baseline'' humans using the extended BDI models described above, representations
of humans augmented with AI tools, and autonomous AI agents, all interacting in social situations of interest, such as elections or epidemics.

This blue sky idea also introduces a set of challenges for ABMs and simulation, which we discuss below. To start with, we note that ABMs are 
also generative models that produce complete and richly structured large-scale data sets. However, the theoretical and methodological challenges 
they pose are different from generative models in machine learning.

\smallskip
\noindent
\textbf{Realistic simulation design:} For answering specific questions, it is good to have the specifics represented in the model. For 
example, to assess the impact of school closures on mitigating an epidemic, it is good to have a model that represents schools well, and
also represents the activity patterns of children when schools are closed. Designing realistic simulations of the problems listed in
Section~\ref{sec:problem}~is going to be quite challenging in terms of representing the population accurately with demographics, behaviors,
information flow, etc. The use of synthetic populations or digital twins might be appropriate in this regard~\citep{Raghunathan2021synthetic}.

\smallskip
\noindent
\textbf{Scaling:} The ABMs and simulations we are proposing would be large and complex. Scaling these to large numbers of agents is an 
active area of research, even in the current BDI framework~\cite[e.g.,][]{deMooij2023framework}. Constructing simulations with large 
numbers of LLM-based agents~\cite[e.g.,][]{Fourney2024} and complex interactions is going to be challenging, though there have been some early attempts~\citep{Chopra2024llm}.

\smallskip
\noindent
\textbf{Epistemic uncertainty}: Uncertainty quantification (UQ) methods for ABMs generally attempt to estimate aleatoric uncertainty, which is 
the uncertainty in outcomes due to stochasticity in the design of the ABM. However, we need to have an understanding of epistemic uncertainty, 
especially due to ABMs necessarily being ``stylized'' at some level and available datasets having inherent uncertainties. For example, both 
mobility modeling and temperature data have some resolution, so there is an implied ``uncertainty principle'' in an ABM for estimating heat 
exposure. How will that affect an AI agent that does scheduling for a human to help avoid heat exposure? Consequently, what will be the
impact on the agency of that human? There will also be epistemic uncertainty in the behavior of adversaries.

Once we address these challenges and manage to run these simulations, they will also create new challenges in sense-making from the
results. Some of these challenges are discussed below.

\smallskip
\noindent
\textbf{Explainability}: In ML models, explainability means seeing the trees for the forest, i.e., making sense of the individual components 
(inputs, extracted features) that are together creating the output. In traditional ABMs, explainability means seeing the forest for the trees, 
i.e., they are eminently understandable at the individual level because they are designed at the individual level. The challenge is to explain 
the population-level outcomes observed due to the complex interactions that take place within the ABM. When we include complex
generative AI models in ABMs, we will make them doubly hard to explain, at the level of both the trees and the forest. This will require new 
methods in simulation analytics~\citep{swarup19analytics}.

\smallskip
\noindent
\textbf{Causality}: Large and complex ABMs are an intermediate regime between toy models and (solely)
observation real-world data, in that they produce large and richly structured output datasets, can be run
only a few times, but are complete in the sense that there is no missingness in the data and no unobserved
factors influencing the outcomes. This creates an interesting and challenging domain for causality inference.
New information-theoretic techniques might be useful, but computational challenges remain~\citep{MartinezSanchez2024synergy}.

\smallskip
\noindent
\textbf{Forecasting in agentic systems:} In some cases, particularly competitive or adversarial ones, agents can act in a way to reduce
the predictability of the system. For example, in the stock market, if there is any regularity that can be exploited to make a profit, agents
will do so. However, this will cause prices to adjust precisely in such a way that the regularity disappears. This makes the system appear
largely random, making questions of explainability and causality that much harder. How many of the problems that we are interested in (the ones
described in Section~\ref{sec:problem} and others) have this characteristic is an open question.

\smallskip
\noindent
\textbf{Intervention discovery:} While a simulation can show us what goes wrong in a particular scenario, we generally approach the 
problem of fixing it with a priori ideas about possible interventions. This ignores all the rich informational structure and intelligence
built into the ABM itself. Ideally, we should be able to use that to guide us in discovering possible interventions, which can then be
evaluated using the simulation.

\section{The Path Ahead}
\label{sec:path}

We have laid out an argument for using agency as the lens through which to evaluate the potential harms of generative AI tools to society. We
have also discussed the advancements needed both theoretically and methodologically to be able to use the concept of agency in this way.
Agent-based simulations will put these new models into motion, so that we can evaluate particular scenarios. The path ahead is not sequential;
several of these problems can and should be worked on in parallel.
At the same time, it is worth keeping in mind that
generative AI tools can also be enormously beneficial in multiple ways:

\begin{itemize}
    \item By enabling computing substrates with intuitive, seamless interactions, thus truly democratizing computing-based infrastructure.
    \item By providing information in an easy-to-understand and unbiased way, thus creating a more informed and educated population.
    \item By improving decision-making in complex systems, thus increasing efficiency and reducing waste.
\end{itemize}

Thus, perhaps the best application of the new theory, model, and simulations would be:

\smallskip
\noindent
\textbf{Evaluating better futures:} How can AI systems and humans act collaboratively to augment agency, instead of just adversarially? What new
questions does that pose for a theory of agency?

It is useful to take a step back at this stage and note that such a theory, if we are able to develop it, could also be used to think about
other problems as well. There are other, equally urgent, problems facing the world today, such as climate change, which require acting on a
global scale. Will these ideas extend to the collective agency needed for effective action at that scale?

\section*{Acknowledgments}
This work was supported by the USDA-NIFA/NSF AI Research Institutes program, under award No. 2021–67021–35344, 
by NSF Expeditions in Computing CCF-1918656, and by NASA Applied Sciences Program Grant \#80NSSC22K1048.

\bibliographystyle{plainnat}  
\bibliography{refs}

\end{document}